\documentclass[letterpaper, 10 pt, conference]{ieeeconf}  %

\IEEEoverridecommandlockouts                              %

\usepackage{graphicx}
\usepackage{amsmath}
\usepackage{amssymb}
\usepackage{booktabs}
\usepackage{algorithm}
\usepackage{authblk}
\usepackage{algpseudocode}
\usepackage{listings}
\usepackage[font=footnotesize]{caption}
\usepackage{subcaption}
\usepackage{wrapfig}
\usepackage{xspace, multirow, array}
\usepackage{xcolor}
\usepackage{cite}

\newcommand{\methodabbr}{FuSe\xspace}

\newcommand{\eg}{e.g., }

\newcommand{\ntrajs}{27K}

\renewcommand{\baselinestretch}{0.98}

\newcommand{\website}{\url{https://fuse-model.github.io}}

\usepackage{graphicx}
\usepackage[bookmarks=True]{hyperref}

\hypersetup{
	colorlinks=true,
	linkcolor=orange,
	filecolor=magenta,      
	urlcolor=magenta,
	citecolor=orange,
}
\usepackage{cleveref}

\title{\LARGE \bf
Beyond Sight: Finetuning Generalist Robot Policies with\\ Heterogeneous Sensors via Language Grounding
}

\author{Joshua Jones$^{*}$, Oier Mees$^{*}$, Carmelo Sferrazza$^{*}$, Kyle Stachowicz, Pieter Abbeel, Sergey Levine %
\\
\website
\thanks{$^\ast$Equal contribution, authors listed alphabetically. The authors are members of Berkeley AI Research (BAIR), UC Berkeley, USA. Please email correspondence to the lead authors: {\tt\small \{joshuajones, csferrazza, oier.mees\}@berkeley.edu}.}%
}

\begin{document}
\makeatletter
\let\@oldmaketitle\@maketitle%
\renewcommand{\@maketitle}{\@oldmaketitle%
  \includegraphics[width=\linewidth] {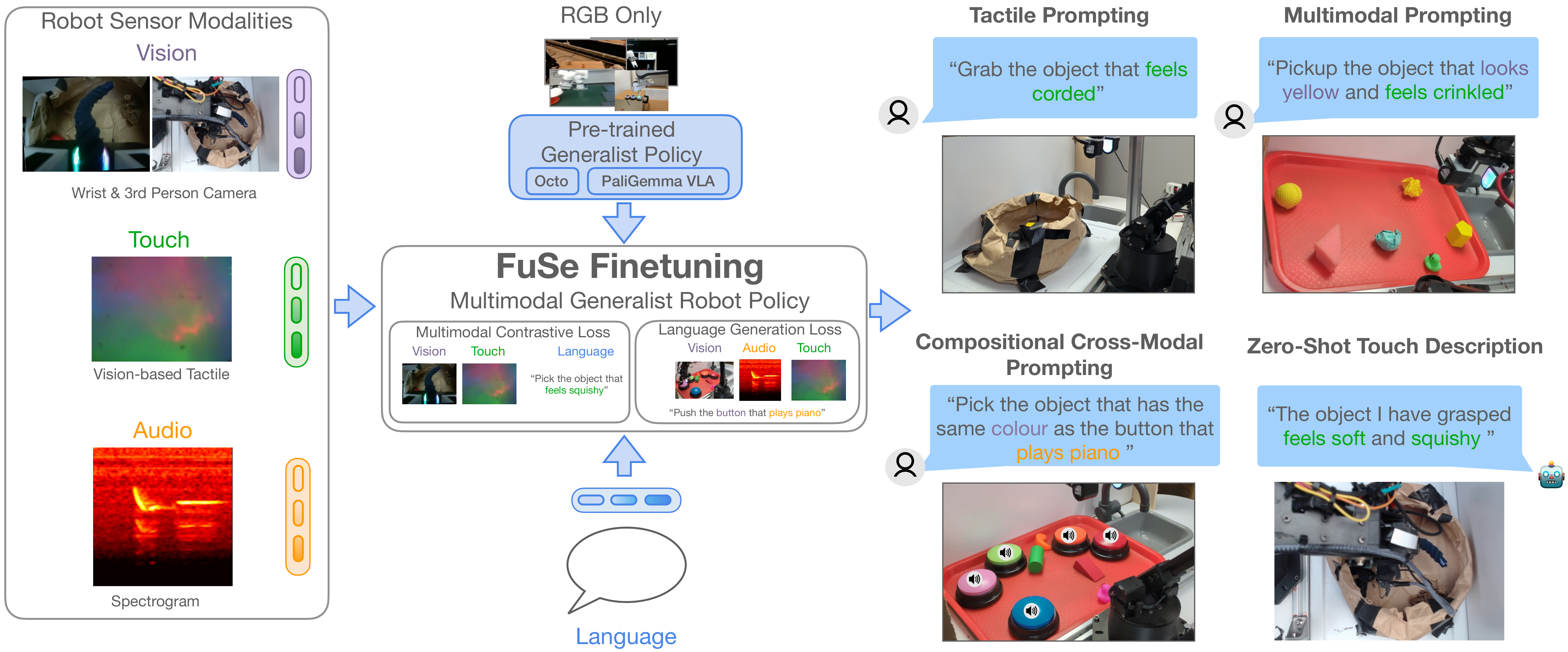} \\[0.25em]
  \refstepcounter{figure}\footnotesize{\bf{Fig. 1:} We introduce \methodabbr, an approach that enables finetuning large image-based pre-trained generalist policies, including vision-language-action (VLA) models, on heterogeneous robot sensor modalities, such as touch or audio, for which large datasets are not readily available, while leveraging natural language as a common cross-modal grounding. Our finetuning recipe enables challenging multimodal and cross-modal prompting tasks in partially-observable scenes and is able to generate zero-shot descriptions of objects it interacts with.}
  \label{fig:real} \medskip \vspace{-10pt}}%
\makeatother

\maketitle

\begin{abstract}

Interacting with the world is a multi-sensory experience: achieving effective general-purpose interaction requires making use of all available modalities -- including vision, touch, and audio -- to fill in gaps from partial observation. For example, when vision is occluded reaching into a bag, a robot should rely on its senses of touch and sound. However, state-of-the-art generalist robot policies are typically trained on large datasets to predict robot actions solely from visual and proprioceptive observations.
In this work, we propose \methodabbr, a novel approach that enables finetuning visuomotor generalist policies on heterogeneous sensor modalities for which large datasets are not readily available by leveraging natural language as a common cross-modal grounding. We combine a multimodal contrastive loss with a sensory-grounded language generation loss to encode high-level semantics. In the context of robot manipulation, we show that \methodabbr~enables performing challenging tasks that require reasoning jointly over modalities such as vision, touch, and sound in a zero-shot setting, such as multimodal prompting, compositional cross-modal prompting, and descriptions of objects it interacts with. We show that the same recipe is applicable to widely different generalist policies, including both diffusion-based generalist policies and large vision-language-action (VLA) models. Extensive experiments in the real world show that \methodabbr~is able to increase success rates by over 20\% compared to all considered baselines.
\end{abstract}

\setcounter{figure}{1}

\section{INTRODUCTION}
Intelligent beings have the ability to seamlessly combine a variety of sensory feedback that allows them to effectively interact with physical the world. Beyond vision, humans rely on the touch and audio feedback to manipulate objects~\cite{johansson2009coding,calandra2018more}, as they provide rich complementary information about object properties, especially when visual information alone might be insufficient to complete the task, such as when locating keys inside a bag~\cite{du2022play}.  
This stands in contrast to state-of-the-art ``generalist'' robot policies~\cite{team2024octo,kim2024openvla,brohan2022rt,brohan2023rt,Doshi24-crossformer} that absorb knowledge from a vast amount of robotics datasets~\cite{open_x_embodiment_rt_x_2023, khazatsky2024droid,walke2023bridgedata,rosete2022corl,mees23hulc2} but typically rely solely on visual and
proprioceptive observations to perform a wide range of tasks.

The main factor limiting development of generalist robot policies based on truly hetereogeneous data is that, while nearly all robotics datasets include visual and proprioceptive information, only a small minority of them include other modalities of sensory data~\cite{fu2024touch, liu2024maniwav, cheng2024towards}. This raises the question: how can we retain the generalization capabilities of generalist robot policies pre-trained on large amounts of data, while connecting their semantic knowledge with heterogeneous sensory data for which large datasets are not readily available?

Prior studies show that natural language can provide a common interface between mixed-modal models, even when they are trained on minimally overlapping data domains~\cite{zeng2022socratic,liang2023code,ahn2022can,huang23vlmaps,huang23avlmaps,driess2023palm,langsurvery24ijrr}. Moreover, relating human language to multimodal percepts and actions naturally enables indexing goals using open-vocabulary queries mixing concepts from multiple distinct modalities (``pick up the \textbf{squishy, red} object''). Nonetheless, incorporating multiple sensing modalities, such as touch or audio, into robotic policies has thus far proved challenging due to data scarcity and in particular a lack of data including joint reasoning over multimodal percepts and low-level robotic actions~\cite{fu2024touch,cheng2024towards,yang2024binding,liu2024maniwav,calandra2018more,calandra2017feeling,du2022play,bi2021zero,wang2020swingbot}.

In this work, we address these challenges and present a recipe to finetune generalist robot policies on smaller-scale datasets comprising modalities complementary to vision, such as touch and sound, and demonstrate that novel capabilities and cross-modal semantic understanding are unlocked through this multimodal finetuning procedure. 

Our key insight is that by grounding all modalities in a single common natural-language modality by way of an auxiliary loss, we can achieve joint reasoning over all modalities. By doing so, we enable our policy to perform challenging manipulation tasks that require reasoning jointly over vision, touch, and sound in a zero-shot setting, enabling multimodal prompting, generation of object descriptions upon interaction, and compositional cross-modal prompting. In practice, our policy can successfully fulfill challenging task instructions, such as ``pick the red object that feels soft and makes a loud sound'', ``describe how the grasped object feels like'', ``pick the object that has the same color as the button that plays piano''.

Our results show that policies leveraging a pre-trained generalist robot policy finetuned on multimodal data consistently outperform baselines finetuned only on vision data, or trained from scratch on heterogeneous sensory data. We find that the same general recipe is applicable to generalist policies with widely different architectures, such as Octo~\cite{team2024octo}, a large transformer-based policy trained on the Open X-Embodiment~\cite{open_x_embodiment_rt_x_2023} (OXE) dataset, and a 3B VLA with a PaliGemma~\cite{beyer2024paligemma} vision-language-model VLM backbone.

For our experiments, we collect a dataset consisting of {\ntrajs} robot trajectories including vision, touch, audio, proprioception, and language instructions on three different real-world robotic manipulation tasks. To the best of our knowledge, this dataset is the first of its kind that also contains robot action data, which is key to perform physically grounded multimodal tasks. We open-source all of our data, code and models to support future research in this area.

\section{RELATED WORK}
\subsection{Generalist Robot Policies}
Generalist robot policies have shown promise of consuming diverse large-scale data to unlock generalization in robotic tasks~\cite{team2024octo,kim2024openvla,brohan2022rt,brohan2023rt,driess2023palm,Doshi24-crossformer,etukuru2024robot}. These policies leverage large robot dataset collections~\cite{open_x_embodiment_rt_x_2023,dasari2019robonet,khazatsky2024droid} that have recently been made available to the community, and are most often queried with language instructions defining the task. In some instances, robot actions are fused with a vision-language model (VLM) backbone~\cite{brohan2023rt,kim2024openvla,driess2023palm,Zawalski24-ecot}, improving generalization due to pre-training on internet-scale data.

However, while some of the recently introduced models~\cite{team2024octo,Doshi24-crossformer} can naturally process flexible observations, the scarcity of datasets that include other sensory modalities, such as touch or audio, limits their capabilities primarily to visual inputs. In contrast, our work shows how such capabilities can be enhanced with a much smaller amount of robotic data containing additional heterogeneous modalities to allow jointly reasoning over modalities, such as vision, touch, and sound in a zero-shot setting.

\subsection{Multimodal Reasoning in Robotics}
Multimodality aims to exploit complementarity across different sensors to enhance the capabilities of autonomous robot policies. Its advantages have repeatedly been shown in the literature, resulting either in improved performance~\cite{calandra2018more, calandra2017feeling, qi2023general, sferrazza2023power, du2022play, mees16iros, mejia2024hearing, li2022see, yuan2024robot, li2019connecting, lee2020making, guzey2024see,du2022play,tian2020contrastive}, generalization~\cite{sferrazza2023power, lin2024learning}, or robustness~\cite{miller1999integration, lee2020making}.

Despite this evidence, only a minority of works employ sensor modalities in addition to vision and proprioception. This is reflected in the robotics datasets made available to the community. For example, the largest collection of robotics dataset, Open X-Embodiment~\cite{open_x_embodiment_rt_x_2023} (OXE), does not include touch or sound as part of their default sensory modalities. Some notable exceptions include recent works~\cite{yu2024octopi,fu2024touch,yang2024binding} that try to align vision, language, and touch for perception tasks. However, most of the available datasets made available through these works do not include robot actions, limiting their applicability for policy training and to perform physically grounded multimodal tasks. Here, we first introduce a multi-task dataset that includes vision, touch, audio, inertial measurements, proprioception, as well as robot actions and language instructions. We then leverage this dataset to finetune large generalist robot models, unlocking novel multimodal reasoning capabilities.

\section{\methodabbr~Finetuning}

\begin{figure*}[t]
\vspace{-2mm}
    \centering
    \includegraphics[width=0.9\textwidth]{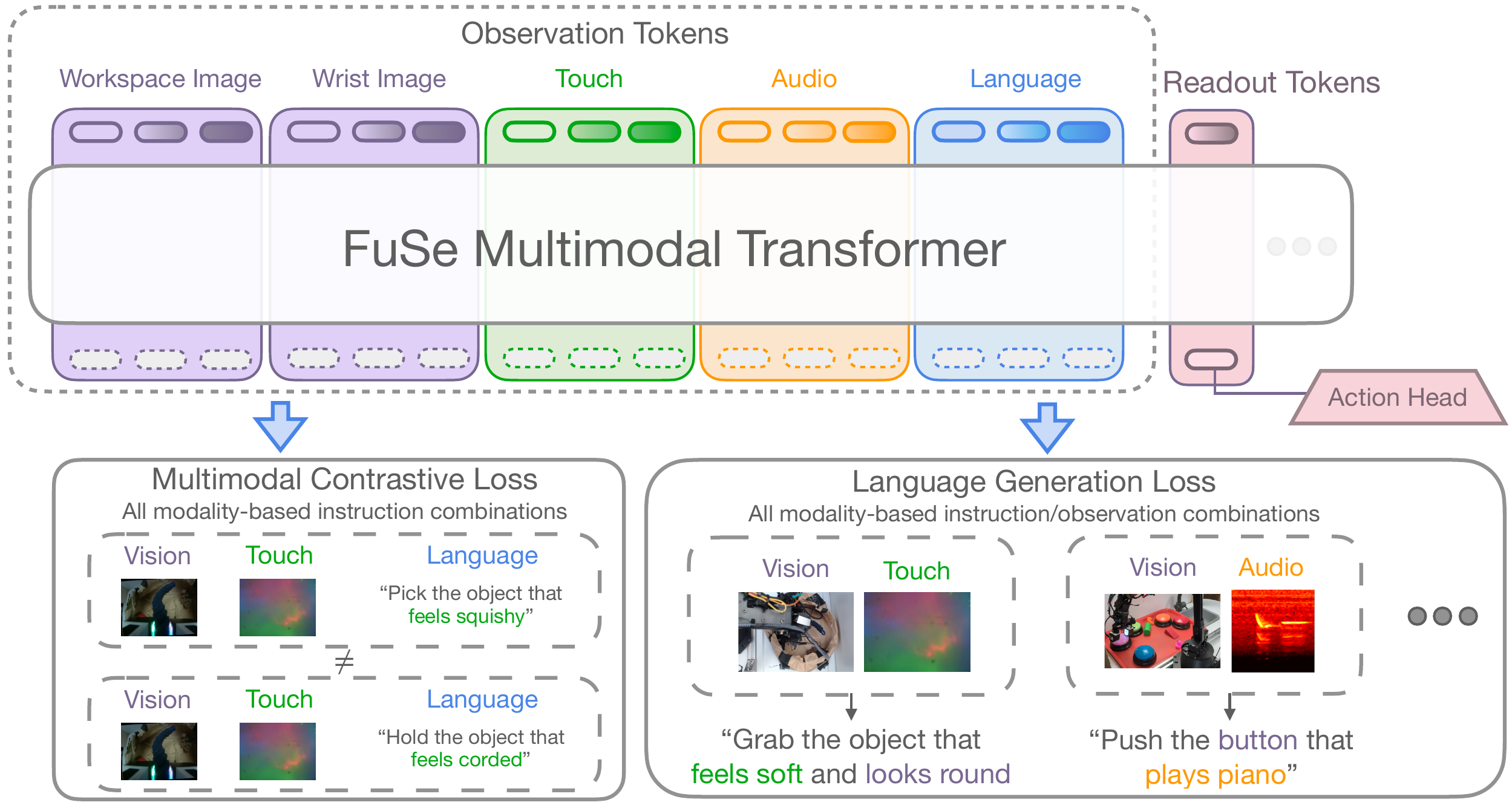}
    \caption{\textbf{Architecture:} We finetune pre-trained generalist robot policies by tokenizing all heteregoneous sensing modalities and passing them though a pre-trained transformer backbone. Crucially, we apply two auxiliary losses that help connect the semantic knowledge of pre-trained generalist policies with new heterogeneous modalities, such as touch and audio. Concretely, we apply both a contrastive loss that aims to maximize mutual information between different views and semantics of the same scene, and a language generation loss that predicts high-level semantics for each modality combination.
    }
    \vspace{-0.6cm}
    \label{fig:arch}
\end{figure*}

State-of-the-art generalist robot policies
typically rely on vision, language, and robot actions as training modalities, which limits their applicability on partially-observable scenes where tasks cannot be completed solely through vision. We propose a recipe, \methodabbr, to \textbf{Fu}se heterogeneous \textbf{Se}nsory data into generalist robot policies. Specifically, we finetune these policies to extend their semantic understanding to include additional sensing modalities, such as touch and sound, while retaining their pre-trained knowledge. By proposing two auxiliary losses, which contrast heterogeneous observations with natural language and generate language from observations, we are able to link a variety of sensing modalities with the semantic knowledge of pre-trained generalist robot policies. We use Octo~\cite{team2024octo}, a transformer-based pre-trained policy, as the backbone model for the main experiments in this paper, but we also show that the same finetuning recipe is applicable to a 3B vision-language-action model based on a PaliGemma~\cite{beyer2024paligemma} VLM backbone. The training architecture is depicted in \Cref{fig:arch}.

This finetuning strategy
introduces three main challenges, namely: (i) the weights of the feature extractors (encoders) for the new modalities generally need to be effectively learned from a small dataset; (ii) the finetuned model empirically tends to predominantly rely on the pre-training modalities, ignoring the new sensors; (iii) novel cross-modal prompting capabilities rely on modality specific annotations, \eg ``the object feels soft and squishy''.
We detail below the modifications required to address all of these challenges.

\textbf{Tactile encoder. } To account for the small finetuning dataset size, we use
a pre-trained tactile encoder and finetune it together with the backbone Octo architecture.
In particular, we use the TVL encoder~\cite{fu2024touch}, which was pre-trained via pairwise contrastive learning across vision, language, and tactile modalities. We feed all tactile images (two in our robot setup) separately through the same TVL encoder. 

\textbf{Audio encoder. } As the raw audio waveform is highly dimensional and noisy, we process the audio data to build a spectrogram as reported in previous work~\cite{guzhov2021esresne,guzhov2022audioclip,gong2021ast,du2022play}. The spectrogram is then treated as a regular image and fed through a ResNet26 encoder~\cite{dosovitskiy2020image}.

\textbf{Auxiliary losses. } As aforementioned, a na\"{\i}ve way of simply finetuning pre-trained generalist policies with a mean-square-error (MSE) imitation loss $\mathcal{L}_{BC}$ conditioned on additional sensor data, leads to the policy over-relying  on its pre-training modalities and ignoring the new modalities.
We overcome this limitation by introducing two additional losses that fully leverage multimodality and connect the semantic knowledge of pre-trained generalist policies with unseen sensor modalities:
\begin{enumerate}
    \item \textit{Multimodal Contrastive Loss: } We introduce a loss that aims to align the various language instructions with the observations via CLIP-style contrastive learning~\cite{radford2021learning}. At a high level, it aims to maximize mutual information between different modalities and semantics of the same scene.
    Concretely, we build an observation embedding by feeding all modalities once more through the transformer and combining them via a multi-head attention layer. We then compute a CLIP-style loss for each possible instruction resulting from combining the different available modalities. These losses are finally averaged to form a combined multimodal contrastive loss $\mathcal{L}_{contrast}$.
    \item \textit{Multimodal Generative Loss: } We design a generative network that functions as an add-on head to the backbone model. In practice, for each possible modality combination, we build an observation embedding as above, and feed it through the generative head. Then, we compute an auxiliary cross-entropy loss $\mathcal{L}_{gen}$ by comparing the head output with the appropriate language instruction. We use a single transformer as the generative head for all possible modality combinations, with modality tokens to distinguish between input modalities. 
\end{enumerate}

The final loss is given by $ \mathcal{L} =  \mathcal{L}_{BC} + \beta  \mathcal{L}_{gen} +  \lambda \mathcal{L}_{contrast}$, where the contrastive loss and the generative loss are summed to the MSE action loss during training.

\textbf{Language Rephrasing. } As discussed previously, cross-modal prompting capabilities require modality specific annotations, e.g., ``the object
feels squishy and looks round''. We annotate the robot trajectories we collect with heterogeneous sensors with after-the-fact language annotations. We annotate these trajectories with templated language that enables us to create augmentations based on multiple sensor inputs,  ``the object
feels squishy and is red'' or ``the object
feels metallic and sounds clinking''. However, at test time we would like users to instruct the policy with free-form language.
To increase the range of possible input instructions, we augment the instructions in the dataset by querying a large language model, ChatGPT~\cite{schulman2022chatgpt}, to generate rephrased versions of the original templates that preserve the original semantic meaning.

\textbf{Implementation Details. } We train all models for 50,000 steps on a v5e-128 TPU pod with a batch size of 1024. We use a cosine learning rate scheduler with 2000 warmup steps, and a peak value of  $3 \times 10^{-4}$. Our language rephrasing buffer contains 20 different templates for each possible modality combination. We set $\beta=1$ and $\lambda=1$ for all experiments.

\section{Experiments}
In this section, we investigate the effectiveness of \methodabbr to finetune pre-trained generalist robot policies to include additional sensor modalities, while linking them to the policy's pre-trained semantic knowledge. We answer the following questions:%

\begin{figure}[t]
    \vspace{-2mm}
    \centering
    \includegraphics[width=0.8\columnwidth]{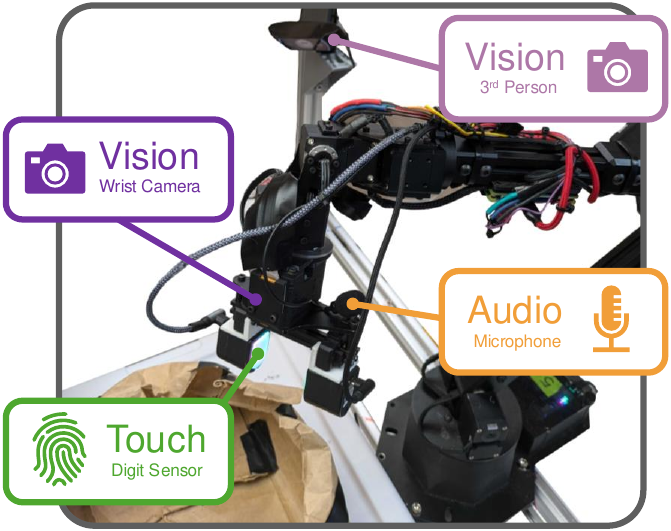}
    \caption{Visualization of the various sensor modalities on our WidowX robot.}
    \label{fig:robot_setup}
    \vspace{-0.5cm}
\end{figure}

\begin{enumerate}
    \item \textbf{Does \methodabbr help perform multimodal prompting tasks in a zero-shot manner in partially observable environments?} (\Cref{sec:finetuning_results})
    \item \textbf{Does \methodabbr~enable multimodal prompting to discriminate between objects that would be ambiguous as described by a single modality?} (\Cref{sec:mm_prompting_results})
    \item \textbf{Can the multimodal capabilities of \methodabbr be applied to compositional reasoning tasks?} (\Cref{sec:compositional_results})
    \item \textbf{Are the proposed auxiliary cross-modal language grounding losses necessary to achieve high performance when finetuning \methodabbr?} (\Cref{sec:ablations_results})
    \item \textbf{Is \methodabbr~ applicable to different generalist robot policy architectures?} (\Cref{sec:vla_results})
\end{enumerate}

\subsection{Real Robot Setup and Training Data}
\begin{figure}[t]
    \vspace{-2mm}
    \centering
    \begin{subfigure}[b]{0.49\linewidth}
        \centering
        \includegraphics[width=\textwidth]{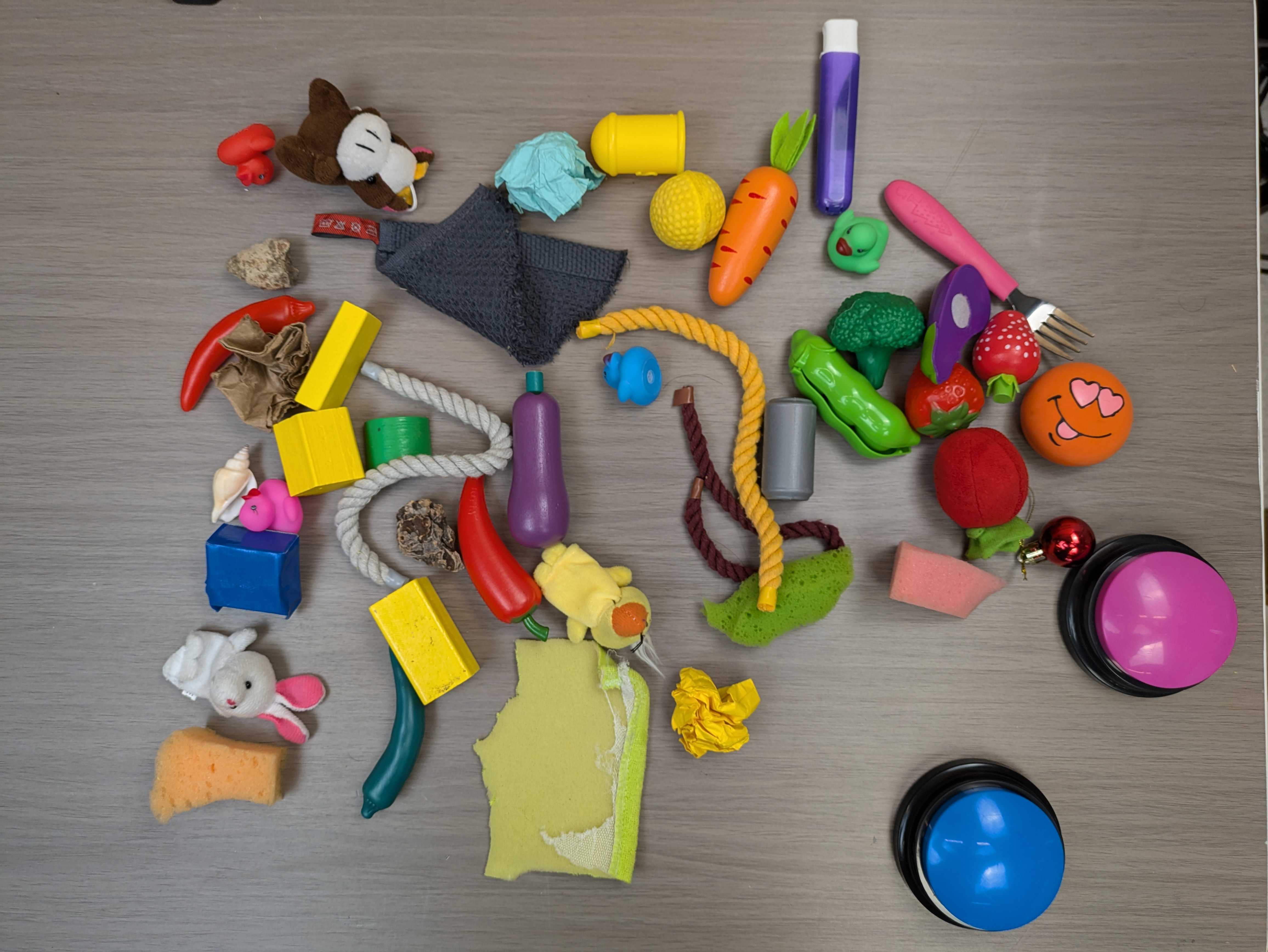}
        \caption{Objects used for evaluation purposes.}
        \label{fig:sub1}
    \end{subfigure}
    \hfill
    \begin{subfigure}[b]{0.49\linewidth}
        \centering
        \includegraphics[width=\textwidth]{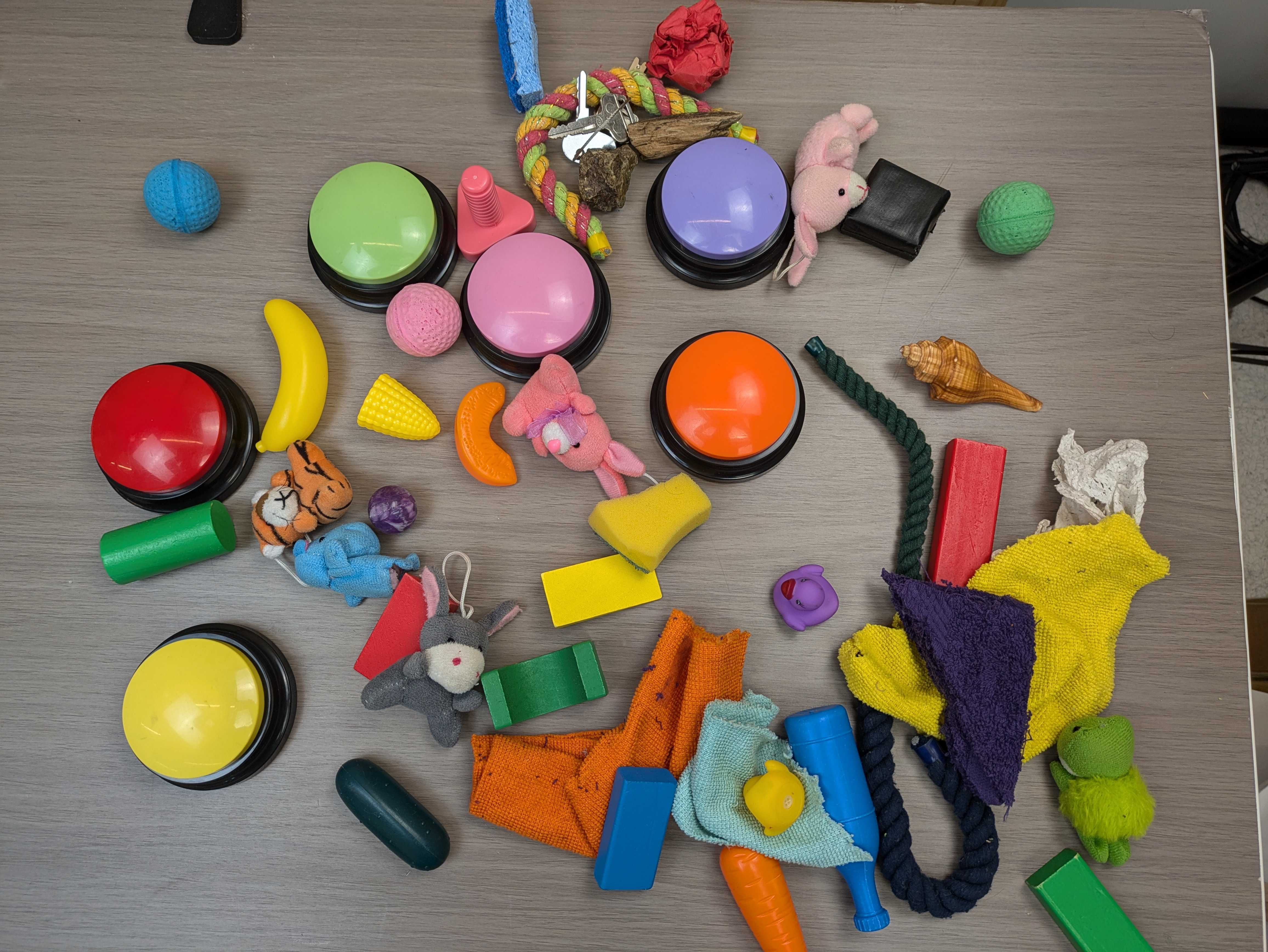}
        \caption{Objects included in the training  data.}
        \label{fig:sub2}
    \end{subfigure}
    \caption{Visualization of objects for real-world experiments, including objects seen (a) and unseen (b) in the multimodal finetuning dataset. Objects differ in shape, appearance, material, hardness, and surface properties.}
    \label{fig:objects}
    \vspace{-0.5cm}
\end{figure}

All our experiments feature a WidowX 250 6-DoF robot arm. The robot is controlled via delta end-effector position commands at a frequency of 5 Hz. The system is equipped with a third-person view RGB camera, a wrist RGB camera, two DIGIT tactile sensors at the gripper fingers, a standard microphone, and a 9-DoF IMU. We present experiments on three different tasks, which are described below. For the grasping scenarios, we evaluate on the 24 objects present in the training dataset, along with 32 unseen test objects; for the button tasks, we evaluate on the six buttons and 13 of the 18 distractors/grasping targets seen in the training dataset, as well as two unseen buttons and 12 unseen distractors.   We visualize the training and test objects used in \Cref{fig:objects}.

We evaluate each model on several different scenarios (e.g., different objects and distractors) for each of the tasks, by running the same scenario for 5 different rollouts.

We collect a dataset of 26,866 trajectories collected via teleoperation using a Meta Quest 2 VR headset. Each trajectory is labeled with a templated language instruction. The two grasping tasks (tabletop and shopping bag) feature visual, tactile, and action data, while the button pressing tasks also includes sound. Visual observations are recorded at a resolution of 640x480, while DIGIT images at a resolution of 320x240. We follow previous work and subtract a static ``background'' image from the tactile observations to emphasize the deviation from the zero-deformation state and reduce systematic differences across DIGIT instances~\cite{calandra2018more}. The audio observations comprise 1s of the most recent microphone samples, recorded at a frequency of 44,100Hz. We visualize our robot sensory setup in Figure ~\ref{fig:robot_setup}.

\begin{figure*}
\vspace{-2mm}
    \centering
    \includegraphics[height=0.04\textwidth]{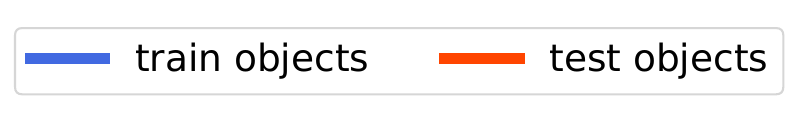} \\
    \includegraphics[height=0.2\textwidth]{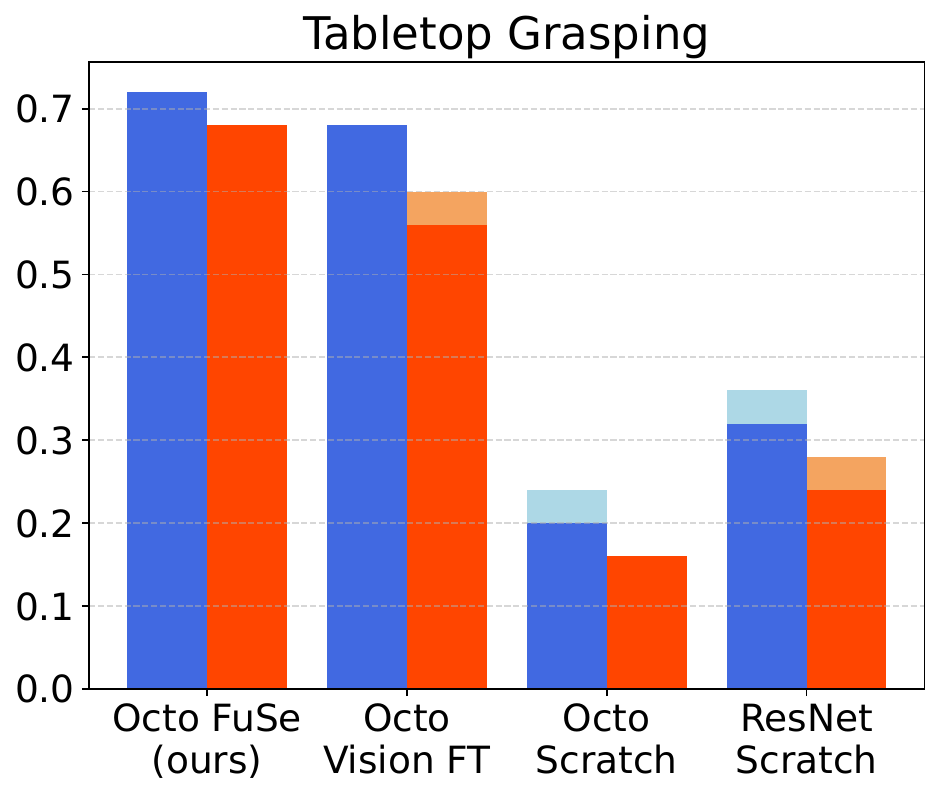}\hspace{-0.5mm}
    \includegraphics[height=0.2\textwidth]{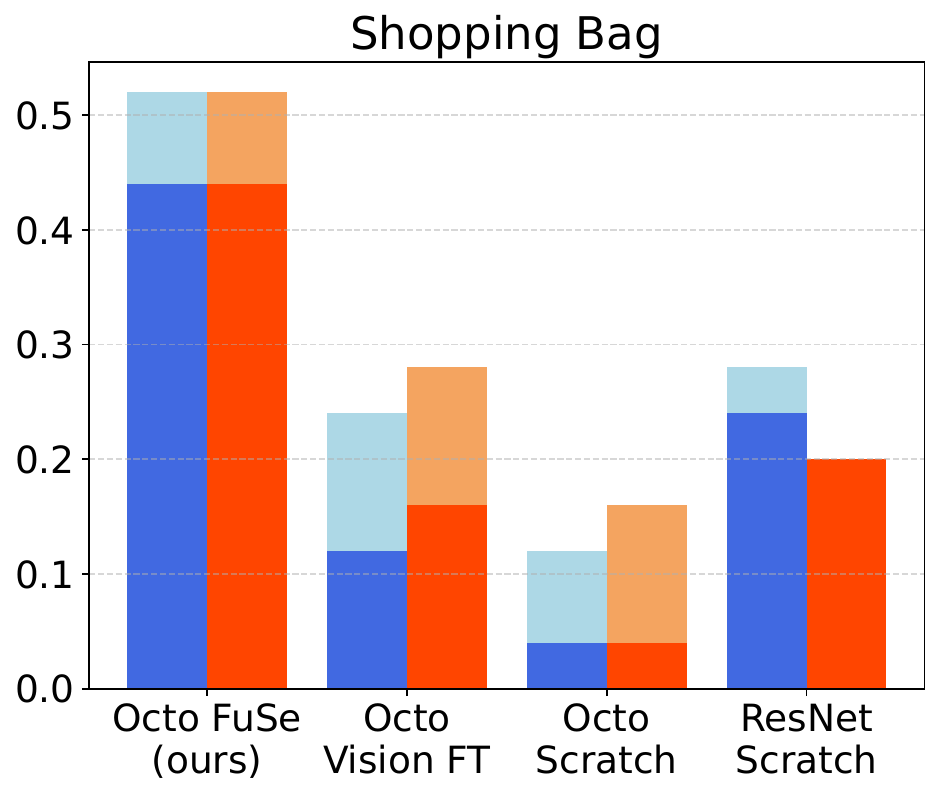}\hspace{-0.5mm}
    \includegraphics[height=0.2\textwidth]{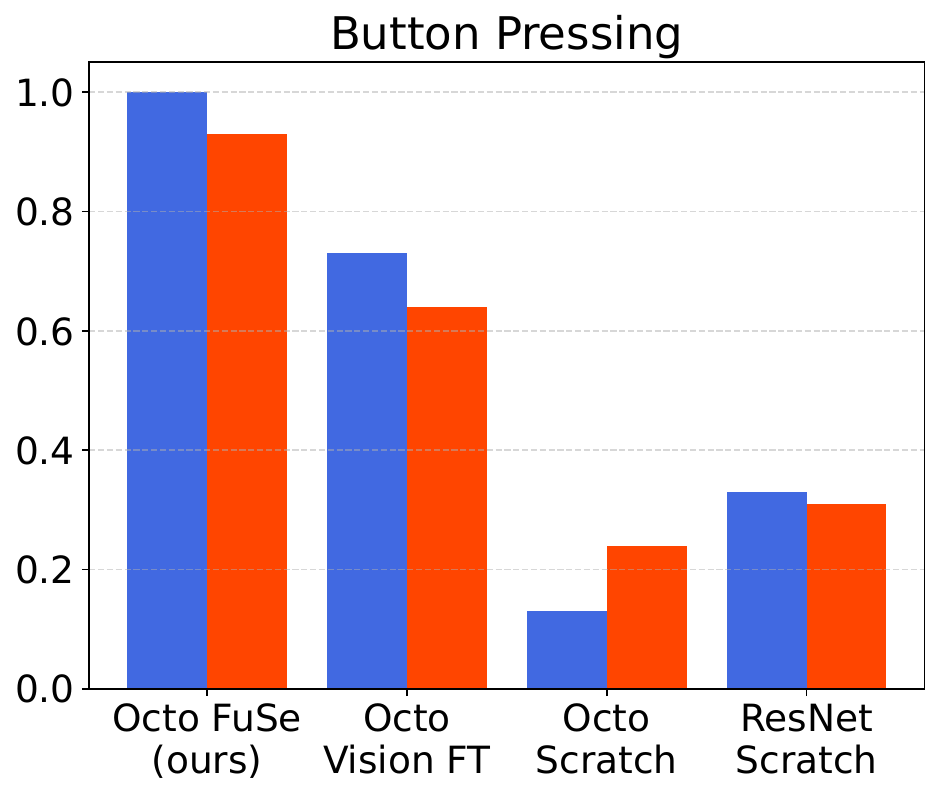}\hspace{-0.5mm}
    \hspace{1mm}\vline\hspace{0.25mm}
    \includegraphics[height=0.2\textwidth]{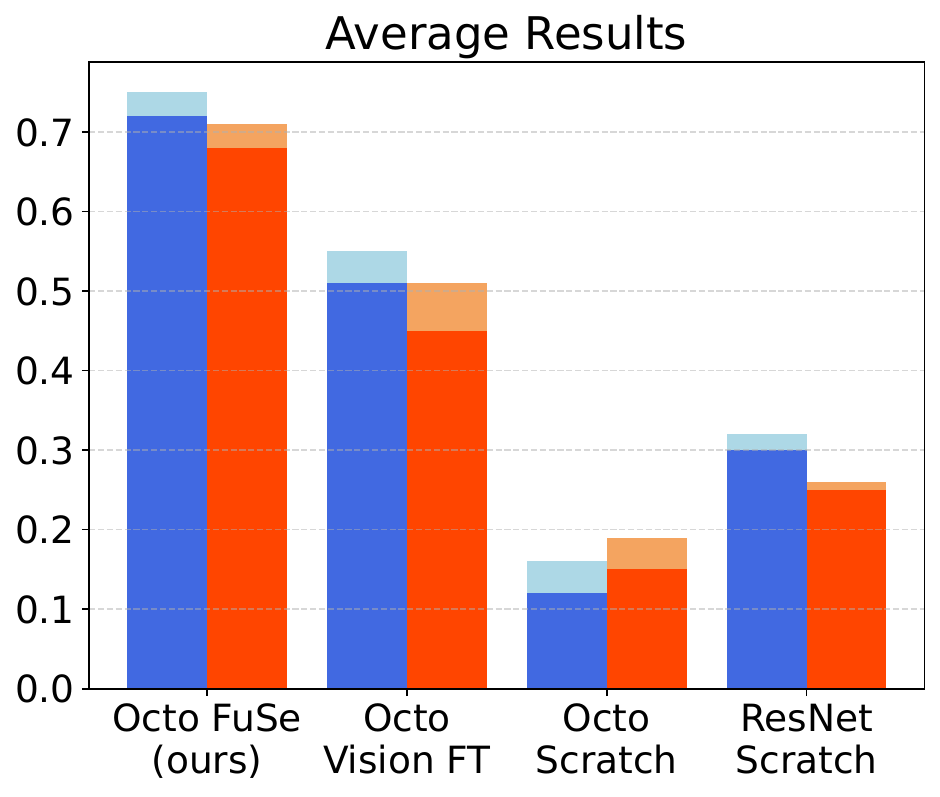}
    \caption{\methodabbr~performance on evaluation tasks compared against baselines. Our approach outperforms baselines trained from scratch or finetuned with vision only, especially on the shopping bag task, which presents partially observable visual scenarios. Lighter shades of color represent intermediate task success, i.e., object touched but not fully grasped.}
    \label{fig:main_results}
    \vspace{-0.6cm}
\end{figure*}

\subsection{Evaluation Tasks}
We design a challenging suite of tasks, which focuses on testing the policies' ability to reason jointly over vision, sound, and touch in a zero-shot setting:

\noindent\textbf{Tabletop Grasping. }
We set up a simple tabletop grasping scenario, where multiple objects are placed on a tray and the task is to grasp the right object as prompted via a text instruction (e.g., pick the carrot).

\noindent\textbf{Shopping Bag. }
This environment presents a more complex grasping scenario, where objects are placed inside a paper bag. This scenario generally features occlusions to third-person view camera, as well as results in poor lighting conditions for the wrist camera as soon as the gripper enters the bag. Thus, this represents an environment with partially-observable visual scenarios.

\noindent\textbf{Button Pressing. }
In this environment, we leverage the sound modality, featuring six sound-making buttons, each playing different sounds upon pressure. The goal is to press the right button depending on the prompt, which can present either visual- or audio-related commands (e.g., ``press the red button'', ``press the button that plays piano'', etc.).

We also devise two multimodal compositional reasoning tasks in the button-pressing setting, where the objective is either i) to grasp objects that share visual characteristics with one of the buttons (e.g., ``grab the object that has the same color as the button that plays piano''), or ii) to press among the training buttons the one that plays the same sound as an unseen button (e.g., ``press the button that plays the same sound as the blue button'').

\vspace{-0.5mm}
\subsection{Finetuning Performance}
\label{sec:finetuning_results}
We investigate the benefits of our multimodal finetuning recipe, which initializes the policy with the Octo generalist policy and is pre-trained on the large OXE robotics dataset~\cite{open_x_embodiment_rt_x_2023}. First, we ask whether pre-training is necessary by comparing our model's performance to a model with the same architecture, but trained from scratch. The results in \Cref{fig:main_results} show a large gap between the models, indicating that training Octo from scratch on our multimodal dataset without our finetuning recipe is challenging due to the limited size of the dataset. In contrast, our approach leverages the knowledge acquired during pre-training and can adapt to the new tasks and modalities with a smaller amount of additional data. Finally, we compare against a ResNet-based baseline, where language instructions are fed through FiLM conditioning~\cite{perez2018film} as in \cite{jang2022bc}. The smaller ResNet26 performs slightly better than training Octo from scratch, but still significantly underperforms our model on all three tasks.

To validate the effect of the new modalities on finetuning performance, we compare with a recipe that finetunes Octo only using the available pre-trained modalities, i.e., vision and action. The results in \Cref{fig:main_results} show how this baseline is competitive on the simpler tasks (tabletop and button pressing), but it is considerably inferior to our model on the bag task, where visual occlusions make visual features less discriminative when the gripper enters the shopping bag.

\subsection{Multimodal Prompting}
\label{sec:mm_prompting_results}

In addition to improving finetuning performance, our training recipe provides the model with additional multimodal capabilities, such as the possibility to provide a multimodal prompt that can successfully discriminate objects based not only on visual features but also based on other modalities such as touch or sound. The evaluation prompts contain several instances where the task is to grab an object with an ambiguous description for one modality, but unique for another (e.g., ``grab the round object that feels squishy'', where the scene presents both a foam ball and a crumpled paper ball). The results are shown in \Cref{fig:multimodal_prompting} for the grasping tasks, on scenarios that present objects sharing the same visual and tactile features, respectively. This experiment demonstrates that our policy can incorporate multimodal instructions to improve over ambiguous descriptions.

\begin{table}[b]
\vspace{-2mm}
\centering
\bgroup
\subfloat[][Vision-ambiguous objects]{
\begin{tabular}{c|cc|cc|}
         & \multicolumn{2}{c|}{Visual} & \multicolumn{2}{c|}{Visual, Tactile} \\ \cline{2-5} 
         & Reach        & Grasp        & Reach            & Grasp            \\ \hline
Tabletop & 0.43         & 0.43         & 0.5              & 0.43             \\ \hline
Bag      & 0.3          & 0.25         & 0.55             & 0.3              \\ \hline
Average  & 0.37         & 0.34         & 0.53             & 0.37             \\ \hline
\end{tabular}
}
\vspace{3mm}
\subfloat[][Touch-ambiguous objects]{
\begin{tabular}{c|cc|cc|}
         & \multicolumn{2}{c|}{Tactile} & \multicolumn{2}{c|}{Visual, Tactile} \\ \cline{2-5} 
         & Reach         & Grasp        & Reach            & Grasp            \\ \hline
Tabletop & 0.4           & 0.4          & 0.4              & 0.4              \\ \hline
Bag      & 0.35          & 0.3          & 0.5              & 0.3              \\ \hline
Average  & 0.38          & 0.35         & 0.45             & 0.35             \\ \hline
\end{tabular}
}

\egroup
    \caption{Multimodal prompting results obtained with the \methodabbr~policy on scenarios that present objects sharing the same visual (a) or tactile (b) features. Our policy incorporate multimodal instructions and improves over ambiguous descriptions.}
    \label{fig:multimodal_prompting}
\end{table}

\vspace{-0.5mm}
\subsection{Compositional Capabilities}
\label{sec:compositional_results}
\begin{figure}
\vspace{-2mm}
    \centering
    \includegraphics[width=0.23\textwidth]{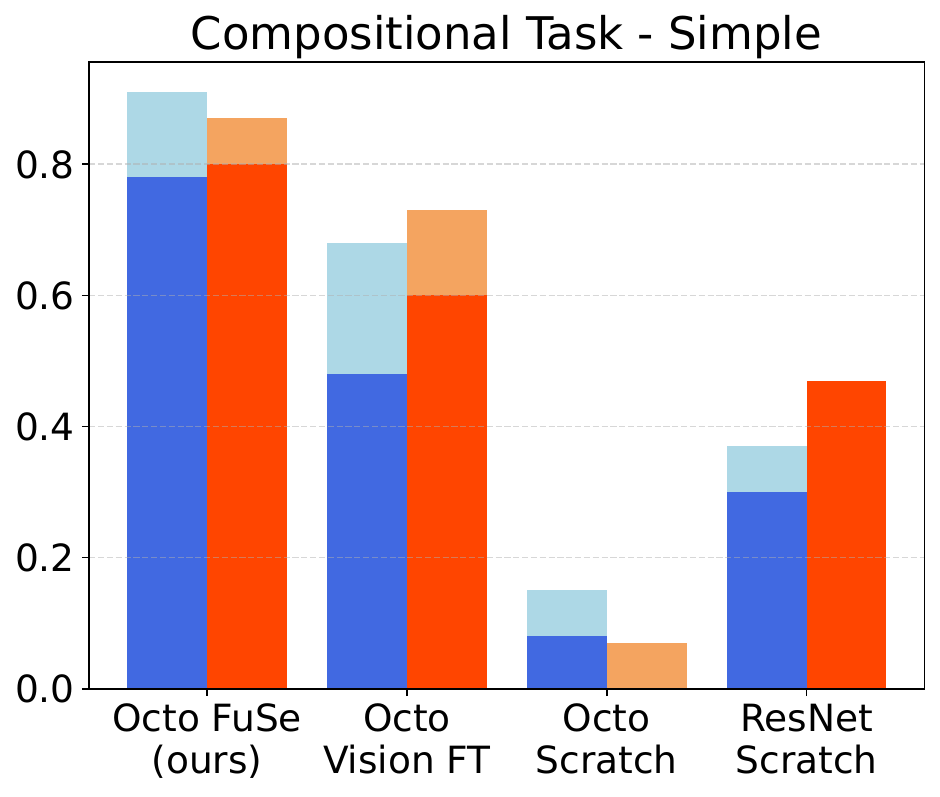}
    \includegraphics[width=0.23\textwidth]{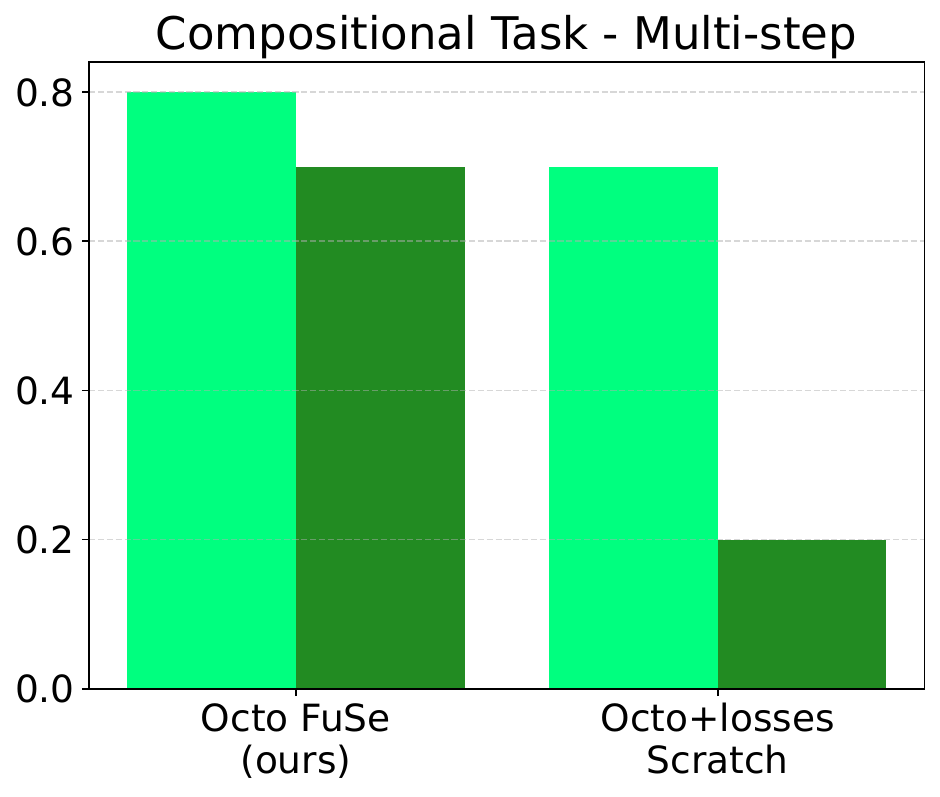}
    \includegraphics[height=0.04\textwidth]{figures/legend.pdf} \includegraphics[height=0.04\textwidth]{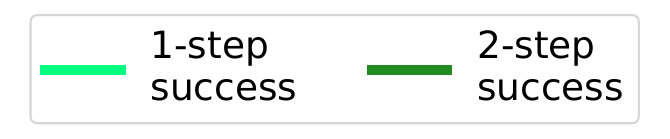}
    \caption{Results on the compositional tasks devised in the button pressing environment. On the left, the instructions are of the type ``pick the \textit{object} that has the same color as the button that play \textit{piano}''. On the right, the whole multi-step task is represented by an instruction like ``press the train button that plays the same sound as the \textit{blue} button''.}
    \vspace{-6mm}\label{fig:compositional_tasks}
\end{figure}

Finally, we showcase compositional capabilities of our model with two different compositional tasks in the button pressing environment:
\begin{itemize}
    \item In a simpler task, we prompt the model to grab an object that has the same color as the training button the plays a certain sound (e.g., ``grab the object with the same color as the button that plays piano'').
    \item In a multi-step task, we exploit the generative head to connect between different subtasks. First, we prompt the model to press a button not seen at training time, using only visual instructions (e.g., ``press the blue button''). Then, we feed the resulting sound to the generative head, which will generate the instruction related to the corresponding audio (e.g., ``press the button that plays piano''). Finally, we prompt the model with the audio instruction in the training environment, where the model has already associated the visual cues of the button to the corresponding sound, and will execute a trajectory that ends up pressing the button that plays the same sound as the button pressed in the first subtask.
\end{itemize}
We report quantitative results in \Cref{fig:compositional_tasks}, showing that even on the simple compositional task, \methodabbr~outperforms all baselines, exploiting its multimodal reasoning capabilities. For the multi-step task, we compare with Octo trained from scratch on all available sensors and with the same auxiliary losses. Once again, \methodabbr~outperforms the baseline, particularly on the full task completion. In fact, the model trained from scratch shows poor language grounding and does not succeed in fulfilling the audio-based instruction.

\begin{figure}[b]
    \vspace{-4mm}
    \centering
    \includegraphics[width=0.25\textwidth]{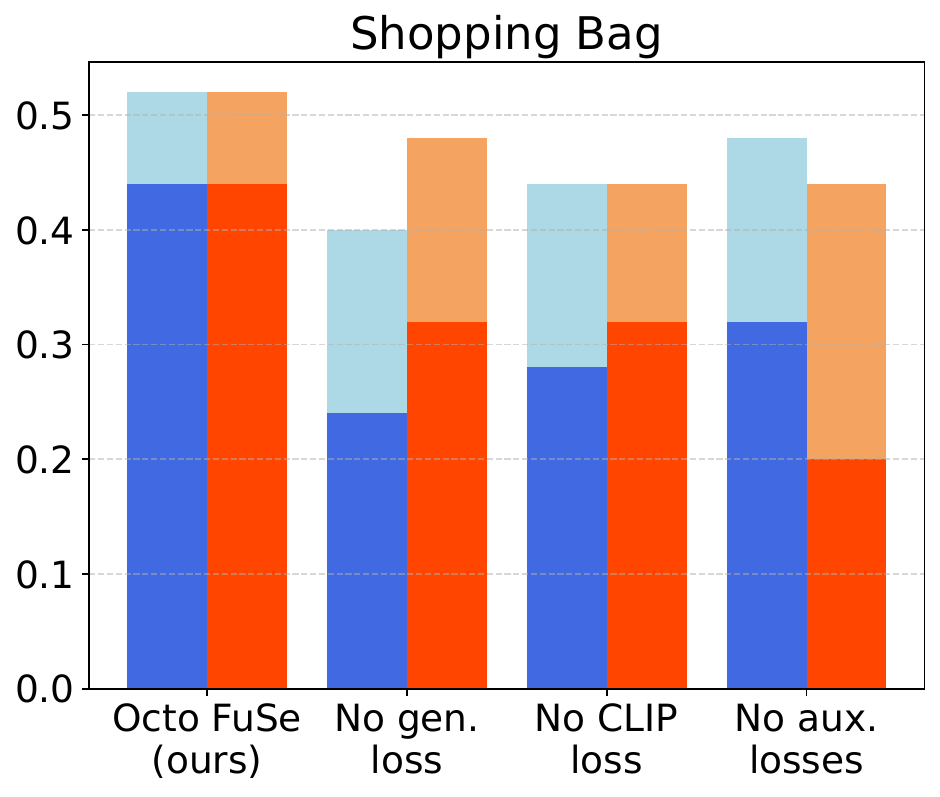}
    \includegraphics[height=0.04\textwidth]{figures/legend.pdf}
    \vspace{-2mm}
    \caption{We study the effect of the proposed losses in an ablation experiment in the shopping bag environment. Our model that includes both contrastive and language generative losses outperforms models trained with only one of the two auxiliary losses or neither.}
    \label{fig:ablation}
\end{figure}
\subsection{Ablation Study}
\label{sec:ablations_results}
We ablate the different \methodabbr~auxiliary losses in the shopping bag task, which features partially observable visual scenarios. \Cref{fig:ablation} shows that including both losses is key to fully exploit the heterogeneous feedback available on the robot, with the performance particularly deteriorating for the baselines on unseen test objects.

\subsection{Vision-Language-Action Model Results}
\label{sec:vla_results}
\begin{figure}
\vspace{-2mm}
    \centering
    \includegraphics[width=0.155\textwidth]{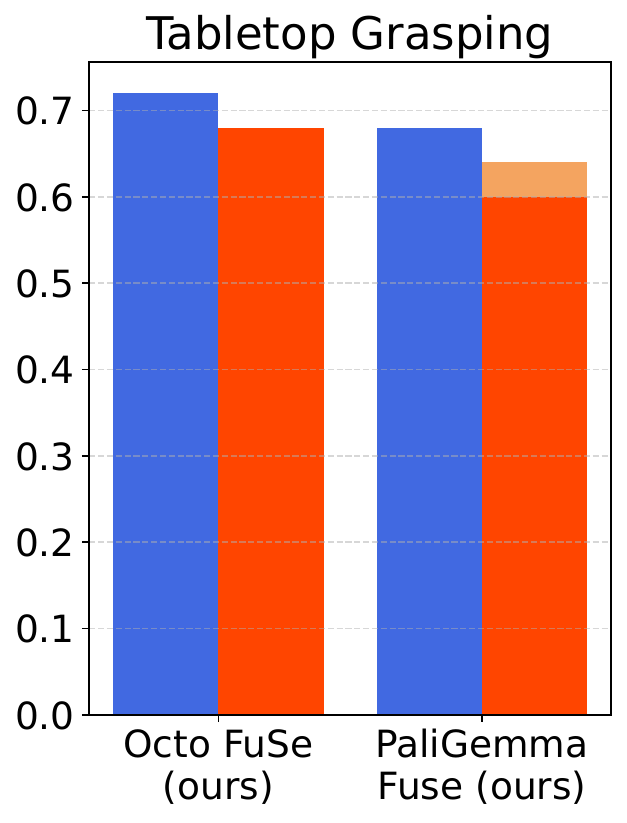}
    \includegraphics[width=0.155\textwidth]{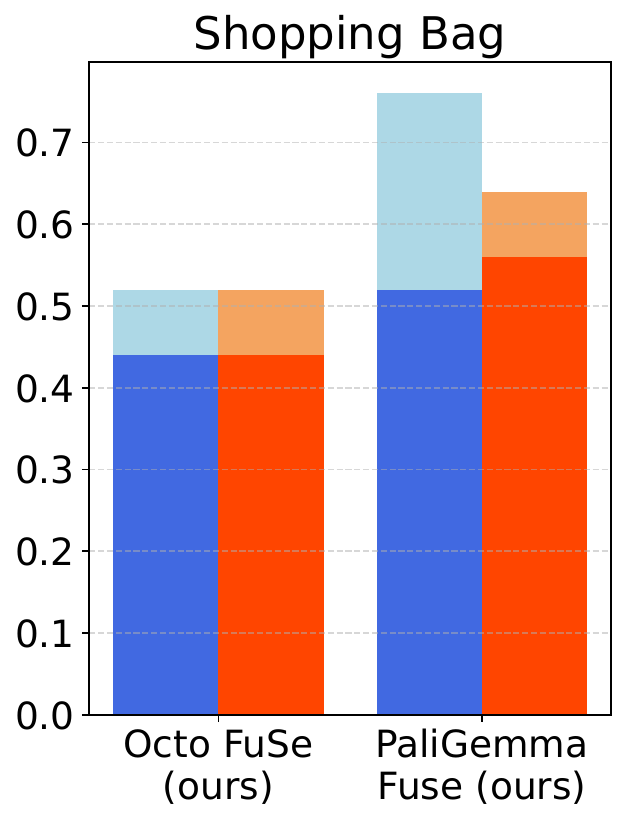}
    \includegraphics[width=0.155\textwidth]{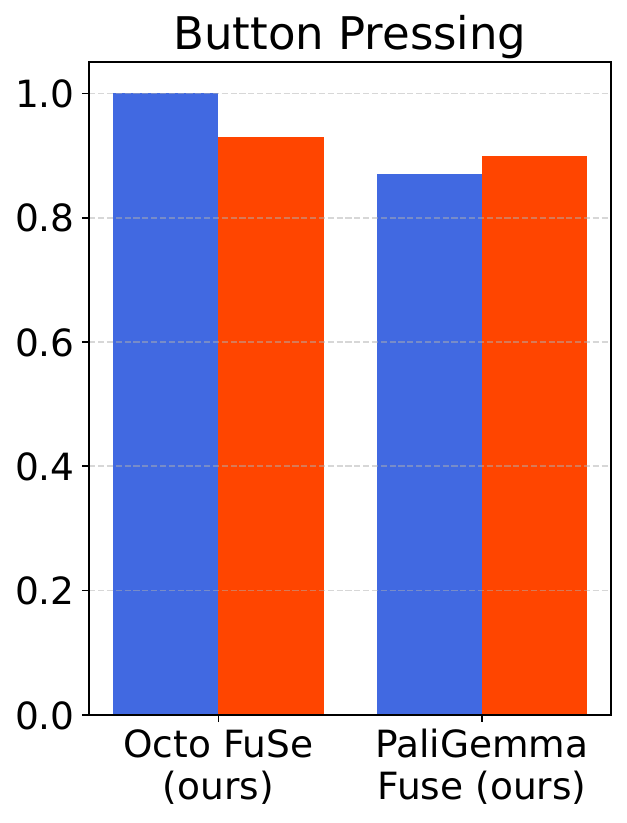}
    \includegraphics[height=0.04\textwidth]{figures/legend.pdf}
    \vspace{-1mm}
    \caption{Performance of a PaliGemma FuSe 3B parameter VLA, trained on our multimodal dataset. Our policy achieves robust performance, showcasing the applicability of FuSe to widely different generalist policies.}
    \label{fig:vla_result}
    \vspace{-0.6cm}
\end{figure}

We also investigate the effectiveness of \methodabbr~to finetune alternative generalist policies based on off-the-shelf \textit{vision-language-action} (VLA) models. Instead of Octo, we finetune a 3B parameter vision-language model to get a VLA model capable of producing both robot actions and language grounding. We use the PaliGemma~\cite{beyer2024paligemma} VLM as the backbone, but modify it to easily incorporate arbitrary observation modalities in a similar fashion to Octo (but unlike other VLA models like OpenVLA~\cite{kim2024openvla}). Such models are also able to incorporate \methodabbr's generative language modeling loss directly rather than requiring an additional language model head, unifying the implementation of action prediction and language-based feature learning. We first pre-train on the OXE dataset with only visual modalities~\cite{palivla}, and finetune on our dataset with all sensor modalities. We note that OXE pre-training generally improves low-level grasping compared to using the original PaliGemma's weights. However, overtraining on OXE during the pre-training phrase appears to harm the policy's language understanding. We choose a checkpoint that pretrains on the OXE dataset for 50,000 steps before finetuning on our multimodal dataset. We show results in \Cref{fig:vla_result}, where the VLA \methodabbr policy is competitive with its Octo-based counterpart and outperforms it on the challenging shopping bag task, demonstrating the effectiveness of \methodabbr across different policy backbones.

To our knowledge, VLA \methodabbr is the first open-source VLA finetuned on heterogeneous (non-visual) sensory inputs.

\section{CONCLUSIONS}
\vspace{-1mm}
In this paper, we introduced \methodabbr, an approach to finetune large, pre-trained robot policies on heterogeneous robot sensor modalities, such as touch or audio, for which large datasets are not readily available. By leveraging natural language as a common cross-modal grounding during training, \methodabbr enables performing challenging tasks that require reasoning jointly over modalities, such as vision, touch, and sound in a
zero-shot setting. \methodabbr enables capabilities such as multimodal prompting, compositional
cross-modal prompting, and descriptions of objects it interacts
with. We also demonstrate the effectiveness of our recipe (multimodal finetuning and feature learning via cross-modal language grounding) is applicable to widely different
generalist policies, including a transformer-based Octo model or a policy finetuned from a generative VLM base model pre-trained on internet-scale data as well as unimodal robot data. 

A limitation of our approach is that training a policy with additional modalities requires increasing training resources, which currently limits our observation history to 0.4s. Increasing training efficiency would enable training with longer context length, potentially leading to improved reasoning about sparse signals such as tactile data, and will be subject of future work.

\section*{ACKNOWLEDGMENT}
This work was supported in part by the ONR Science of Autonomy Program N000142212121, the SNSF Postdoc Mobility Fellowship 211086, the BAIR Industrial Consortium, ONR N00014-20-1-2383 and the AI Institute. Pieter Abbeel holds concurrent appointments as a Professor at UC Berkeley and as an Amazon Scholar. This paper describes work performed at UC Berkeley and is not associated with Amazon.

\bibliographystyle{IEEEtran}
\bibliography{references}

\begin{thebibliography}{10}
\providecommand{\url}[1]{#1}
\csname url@samestyle\endcsname
\providecommand{\newblock}{\relax}
\providecommand{\bibinfo}[2]{#2}
\providecommand{\BIBentrySTDinterwordspacing}{\spaceskip=0pt\relax}
\providecommand{\BIBentryALTinterwordstretchfactor}{4}
\providecommand{\BIBentryALTinterwordspacing}{\spaceskip=\fontdimen2\font plus
\BIBentryALTinterwordstretchfactor\fontdimen3\font minus \fontdimen4\font\relax}
\providecommand{\BIBforeignlanguage}[2]{{%
\expandafter\ifx\csname l@#1\endcsname\relax
\typeout{** WARNING: IEEEtran.bst: No hyphenation pattern has been}%
\typeout{** loaded for the language `#1'. Using the pattern for}%
\typeout{** the default language instead.}%
\else
\language=\csname l@#1\endcsname
\fi
#2}}
\providecommand{\BIBdecl}{\relax}
\BIBdecl

\bibitem{johansson2009coding}
R.~S. Johansson and J.~R. Flanagan, ``Coding and use of tactile signals from the fingertips in object manipulation tasks,'' \emph{Nature Reviews Neuroscience}, vol.~10, no.~5, pp. 345--359, 2009.

\bibitem{calandra2018more}
R.~Calandra, A.~Owens, D.~Jayaraman, J.~Lin, W.~Yuan, J.~Malik, E.~H. Adelson, and S.~Levine, ``More than a feeling: Learning to grasp and regrasp using vision and touch,'' \emph{IEEE Robotics and Automation Letters}, vol.~3, no.~4, pp. 3300--3307, 2018.

\bibitem{du2022play}
M.~Du, O.~Y. Lee, S.~Nair, and C.~Finn, ``Play it by ear: Learning skills amidst occlusion through audio-visual imitation learning,'' \emph{arXiv preprint arXiv:2205.14850}, 2022.

\bibitem{team2024octo}
{Octo Model Team}, D.~Ghosh, H.~Walke, K.~Pertsch, K.~Black, O.~Mees, S.~Dasari, J.~Hejna, C.~Xu, J.~Luo, T.~Kreiman, Y.~Tan, P.~Sanketi, Q.~Vuong, T.~Xiao, D.~Sadigh, C.~Finn, and S.~Levine, ``Octo: An open-source generalist robot policy,'' in \emph{Proceedings of Robotics: Science and Systems}, Delft, Netherlands, 2024.

\bibitem{kim2024openvla}
M.~J. Kim, K.~Pertsch, S.~Karamcheti, T.~Xiao, A.~Balakrishna, S.~Nair, R.~Rafailov, E.~Foster, G.~Lam, P.~Sanketi \emph{et~al.}, ``Openvla: An open-source vision-language-action model,'' \emph{arXiv preprint arXiv:2406.09246}, 2024.

\bibitem{brohan2022rt}
A.~Brohan, N.~Brown, J.~Carbajal, Y.~Chebotar, J.~Dabis, C.~Finn, K.~Gopalakrishnan, K.~Hausman, A.~Herzog, J.~Hsu \emph{et~al.}, ``Rt-1: Robotics transformer for real-world control at scale,'' \emph{arXiv preprint arXiv:2212.06817}, 2022.

\bibitem{brohan2023rt}
A.~Brohan, N.~Brown, J.~Carbajal, Y.~Chebotar, X.~Chen, K.~Choromanski, T.~Ding, D.~Driess, A.~Dubey, C.~Finn \emph{et~al.}, ``Rt-2: Vision-language-action models transfer web knowledge to robotic control,'' \emph{arXiv preprint arXiv:2307.15818}, 2023.

\bibitem{Doshi24-crossformer}
R.~Doshi, H.~Walke, O.~Mees, S.~Dasari, and S.~Levine, ``Scaling cross-embodied learning: One policy for manipulation, navigation, locomotion and aviation,'' in \emph{Conference on Robot Learning}, 2024.

\bibitem{open_x_embodiment_rt_x_2023}
O.~X.-E. Collaboration, A.~Padalkar, A.~Pooley, A.~Jain, A.~Bewley, A.~Herzog, A.~Irpan, A.~Khazatsky, A.~Rai, A.~Singh, A.~Brohan, A.~Raffin, A.~Wahid, B.~Burgess-Limerick, B.~Kim, B.~Schölkopf, B.~Ichter, C.~Lu, C.~Xu, C.~Finn, C.~Xu, C.~Chi, C.~Huang, C.~Chan, C.~Pan, C.~Fu, C.~Devin, D.~Driess, D.~Pathak, D.~Shah, D.~Büchler, D.~Kalashnikov, D.~Sadigh, E.~Johns, F.~Ceola, F.~Xia, F.~Stulp, G.~Zhou, G.~S. Sukhatme, G.~Salhotra, G.~Yan, G.~Schiavi, H.~Su, H.-S. Fang, H.~Shi, H.~B. Amor, H.~I. Christensen, H.~Furuta, H.~Walke, H.~Fang, I.~Mordatch, I.~Radosavovic, I.~Leal, J.~Liang, J.~Kim, J.~Schneider, J.~Hsu, J.~Bohg, J.~Bingham, J.~Wu, J.~Wu, J.~Luo, J.~Gu, J.~Tan, J.~Oh, J.~Malik, J.~Tompson, J.~Yang, J.~J. Lim, J.~Silvério, J.~Han, K.~Rao, K.~Pertsch, K.~Hausman, K.~Go, K.~Gopalakrishnan, K.~Goldberg, K.~Byrne, K.~Oslund, K.~Kawaharazuka, K.~Zhang, K.~Majd, K.~Rana, K.~Srinivasan, L.~Y. Chen, L.~Pinto, L.~Tan, L.~Ott, L.~Lee, M.~Tomizuka, M.~Du, M.~Ahn, M.~Zhang, M.~Ding, M.~K. Srirama, M.~Sharma,
  M.~J. Kim, N.~Kanazawa, N.~Hansen, N.~Heess, N.~J. Joshi, N.~Suenderhauf, N.~D. Palo, N.~M.~M. Shafiullah, O.~Mees, O.~Kroemer, P.~R. Sanketi, P.~Wohlhart, P.~Xu, P.~Sermanet, P.~Sundaresan, Q.~Vuong, R.~Rafailov, R.~Tian, R.~Doshi, R.~Martín-Martín, R.~Mendonca, R.~Shah, R.~Hoque, R.~Julian, S.~Bustamante, S.~Kirmani, S.~Levine, S.~Moore, S.~Bahl, S.~Dass, S.~Song, S.~Xu, S.~Haldar, S.~Adebola, S.~Guist, S.~Nasiriany, S.~Schaal, S.~Welker, S.~Tian, S.~Dasari, S.~Belkhale, T.~Osa, T.~Harada, T.~Matsushima, T.~Xiao, T.~Yu, T.~Ding, T.~Davchev, T.~Z. Zhao, T.~Armstrong, T.~Darrell, V.~Jain, V.~Vanhoucke, W.~Zhan, W.~Zhou, W.~Burgard, X.~Chen, X.~Wang, X.~Zhu, X.~Li, Y.~Lu, Y.~Chebotar, Y.~Zhou, Y.~Zhu, Y.~Xu, Y.~Wang, Y.~Bisk, Y.~Cho, Y.~Lee, Y.~Cui, Y.~hua Wu, Y.~Tang, Y.~Zhu, Y.~Li, Y.~Iwasawa, Y.~Matsuo, Z.~Xu, and Z.~J. Cui, ``Open {X-E}mbodiment: Robotic learning datasets and {RT-X} models,'' in \emph{Proceedings of the IEEE International Conference on Robotics and Automation (ICRA)}, Yokohama, Japan,
  2024.

\bibitem{khazatsky2024droid}
A.~Khazatsky, K.~Pertsch, S.~Nair, A.~Balakrishna, S.~Dasari, S.~Karamcheti, S.~Nasiriany, M.~K. Srirama, L.~Y. Chen, K.~Ellis \emph{et~al.}, ``Droid: A large-scale in-the-wild robot manipulation dataset,'' \emph{arXiv preprint arXiv:2403.12945}, 2024.

\bibitem{walke2023bridgedata}
H.~Walke, K.~Black, A.~Lee, M.~J. Kim, M.~Du, C.~Zheng, T.~Zhao, P.~Hansen-Estruch, Q.~Vuong, A.~He, V.~Myers, K.~Fang, C.~Finn, and S.~Levine, ``Bridgedata v2: A dataset for robot learning at scale,'' in \emph{Conference on Robot Learning (CoRL)}, 2023.

\bibitem{rosete2022corl}
E.~Rosete-Beas, O.~Mees, G.~Kalweit, J.~Boedecker, and W.~Burgard, ``Latent plans for task agnostic offline reinforcement learning,'' in \emph{Proceedings of the 6th Conference on Robot Learning (CoRL)}, Auckland, New Zealand, 2022.

\bibitem{mees23hulc2}
O.~Mees, J.~Borja-Diaz, and W.~Burgard, ``Grounding language with visual affordances over unstructured data,'' in \emph{Proceedings of the IEEE International Conference on Robotics and Automation (ICRA)}, London, UK, 2023.

\bibitem{fu2024touch}
L.~Fu, G.~Datta, H.~Huang, W.~C.-H. Panitch, J.~Drake, J.~Ortiz, M.~Mukadam, M.~Lambeta, R.~Calandra, and K.~Goldberg, ``A touch, vision, and language dataset for multimodal alignment,'' in \emph{Forty-first International Conference on Machine Learning}.

\bibitem{liu2024maniwav}
Z.~Liu, C.~Chi, E.~Cousineau, N.~Kuppuswamy, B.~Burchfiel, and S.~Song, ``Maniwav: Learning robot manipulation from in-the-wild audio-visual data,'' in \emph{8th Annual Conference on Robot Learning}.

\bibitem{cheng2024towards}
N.~Cheng, Y.~Li, J.~Gao, B.~Fang, J.~Xu, and W.~Han, ``Towards comprehensive multimodal perception: Introducing the touch-language-vision dataset,'' \emph{arXiv preprint arXiv:2403.09813}, 2024.

\bibitem{zeng2022socratic}
A.~Zeng, M.~Attarian, B.~Ichter, K.~Choromanski, A.~Wong, S.~Welker, F.~Tombari, A.~Purohit, M.~Ryoo, V.~Sindhwani \emph{et~al.}, ``Socratic models: Composing zero-shot multimodal reasoning with language,'' \emph{arXiv preprint arXiv:2204.00598}, 2022.

\bibitem{liang2023code}
J.~Liang, W.~Huang, F.~Xia, P.~Xu, K.~Hausman, B.~Ichter, P.~Florence, and A.~Zeng, ``Code as policies: Language model programs for embodied control,'' in \emph{2023 IEEE International Conference on Robotics and Automation (ICRA)}.\hskip 1em plus 0.5em minus 0.4em\relax IEEE, 2023, pp. 9493--9500.

\bibitem{ahn2022can}
M.~Ahn, A.~Brohan, N.~Brown, Y.~Chebotar, O.~Cortes, B.~David, C.~Finn, C.~Fu, K.~Gopalakrishnan, K.~Hausman \emph{et~al.}, ``Do as i can, not as i say: Grounding language in robotic affordances,'' \emph{arXiv preprint arXiv:2204.01691}, 2022.

\bibitem{huang23vlmaps}
C.~Huang, O.~Mees, A.~Zeng, and W.~Burgard, ``Visual language maps for robot navigation,'' in \emph{Proceedings of the IEEE International Conference on Robotics and Automation (ICRA)}, London, UK, 2023.

\bibitem{huang23avlmaps}
------, ``Audio visual language maps for robot navigation,'' in \emph{Proceedings of the International Symposium on Experimental Robotics (ISER)}, Chiang Mai, Thailand, 2023.

\bibitem{driess2023palm}
D.~Driess, F.~Xia, M.~S. Sajjadi, C.~Lynch, A.~Chowdhery, B.~Ichter, A.~Wahid, J.~Tompson, Q.~Vuong, T.~Yu \emph{et~al.}, ``Palm-e: An embodied multimodal language model,'' \emph{arXiv preprint arXiv:2303.03378}, 2023.

\bibitem{langsurvery24ijrr}
H.~Zhou, X.~Yao, O.~Mees, Y.~Meng, T.~Xiao, Y.~Bisk, J.~Oh, E.~Johns, M.~Shridhar, D.~Shah, J.~Thomason, K.~Huang, J.~Chai, Z.~Bing, and A.~Knoll, ``Bridging language and action: A survey of language-based robot manipulation,'' \emph{arXiv preprint arXiv:2312.10807}, 2024.

\bibitem{yang2024binding}
F.~Yang, C.~Feng, Z.~Chen, H.~Park, D.~Wang, Y.~Dou, Z.~Zeng, X.~Chen, R.~Gangopadhyay, A.~Owens \emph{et~al.}, ``Binding touch to everything: Learning unified multimodal tactile representations,'' in \emph{Proceedings of the IEEE/CVF Conference on Computer Vision and Pattern Recognition}, 2024, pp. 26\,340--26\,353.

\bibitem{calandra2017feeling}
R.~Calandra, A.~Owens, M.~Upadhyaya, W.~Yuan, J.~Lin, E.~H. Adelson, and S.~Levine, ``The feeling of success: Does touch sensing help predict grasp outcomes?'' in \emph{Conference on Robot Learning}.\hskip 1em plus 0.5em minus 0.4em\relax PMLR, 2017, pp. 314--323.

\bibitem{bi2021zero}
T.~Bi, C.~Sferrazza, and R.~D’Andrea, ``Zero-shot sim-to-real transfer of tactile control policies for aggressive swing-up manipulation,'' \emph{IEEE Robotics and Automation Letters}, vol.~6, no.~3, pp. 5761--5768, 2021.

\bibitem{wang2020swingbot}
C.~Wang, S.~Wang, B.~Romero, F.~Veiga, and E.~Adelson, ``Swingbot: Learning physical features from in-hand tactile exploration for dynamic swing-up manipulation,'' in \emph{2020 IEEE/RSJ International Conference on Intelligent Robots and Systems (IROS)}.\hskip 1em plus 0.5em minus 0.4em\relax IEEE, 2020, pp. 5633--5640.

\bibitem{beyer2024paligemma}
L.~Beyer, A.~Steiner, A.~S. Pinto, A.~Kolesnikov, X.~Wang, D.~Salz, M.~Neumann, I.~Alabdulmohsin, M.~Tschannen, E.~Bugliarello \emph{et~al.}, ``Paligemma: A versatile 3b vlm for transfer,'' \emph{arXiv preprint arXiv:2407.07726}, 2024.

\bibitem{etukuru2024robot}
H.~Etukuru, N.~Naka, Z.~Hu, S.~Lee, J.~Mehu, A.~Edsinger, C.~Paxton, S.~Chintala, L.~Pinto, and N.~M.~M. Shafiullah, ``Robot utility models: General policies for zero-shot deployment in new environments,'' \emph{arXiv preprint arXiv:2409.05865}, 2024.

\bibitem{dasari2019robonet}
S.~Dasari, F.~Ebert, S.~Tian, S.~Nair, B.~Bucher, K.~Schmeckpeper, S.~Singh, S.~Levine, and C.~Finn, ``Robonet: Large-scale multi-robot learning,'' \emph{arXiv preprint arXiv:1910.11215}, 2019.

\bibitem{Zawalski24-ecot}
M.~Zawalski, W.~Chen, K.~Pertsch, O.~Mees, C.~Finn, and S.~Levine, ``Robotic control via embodied chain-of-thought reasoning,'' in \emph{Conference on Robot Learning}, 2024.

\bibitem{qi2023general}
H.~Qi, B.~Yi, S.~Suresh, M.~Lambeta, Y.~Ma, R.~Calandra, and J.~Malik, ``General in-hand object rotation with vision and touch,'' in \emph{Conference on Robot Learning}.\hskip 1em plus 0.5em minus 0.4em\relax PMLR, 2023, pp. 2549--2564.

\bibitem{sferrazza2023power}
C.~Sferrazza, Y.~Seo, H.~Liu, Y.~Lee, and P.~Abbeel, ``The power of the senses: Generalizable manipulation from vision and touch through masked multimodal learning,'' \emph{arXiv preprint arXiv:2311.00924}, 2023.

\bibitem{mees16iros}
O.~Mees, A.~Eitel, and W.~Burgard, ``Choosing smartly: Adaptive multimodal fusion for object detection in changing environments,'' in \emph{Proceedings of the International Conference on Intelligent Robots and Systems (IROS)}, Daejeon, South Korea, 2016.

\bibitem{mejia2024hearing}
J.~Mejia, V.~Dean, T.~Hellebrekers, and A.~Gupta, ``Hearing touch: Audio-visual pretraining for contact-rich manipulation,'' \emph{arXiv preprint arXiv:2405.08576}, 2024.

\bibitem{li2022see}
H.~Li, Y.~Zhang, J.~Zhu, S.~Wang, M.~A. Lee, H.~Xu, E.~Adelson, L.~Fei-Fei, R.~Gao, and J.~Wu, ``See, hear, and feel: Smart sensory fusion for robotic manipulation,'' \emph{arXiv preprint arXiv:2212.03858}, 2022.

\bibitem{yuan2024robot}
Y.~Yuan, H.~Che, Y.~Qin, B.~Huang, Z.-H. Yin, K.-W. Lee, Y.~Wu, S.-C. Lim, and X.~Wang, ``Robot synesthesia: In-hand manipulation with visuotactile sensing,'' in \emph{2024 IEEE International Conference on Robotics and Automation (ICRA)}.\hskip 1em plus 0.5em minus 0.4em\relax IEEE, 2024, pp. 6558--6565.

\bibitem{li2019connecting}
Y.~Li, J.-Y. Zhu, R.~Tedrake, and A.~Torralba, ``Connecting touch and vision via cross-modal prediction,'' in \emph{Proceedings of the IEEE/CVF Conference on Computer Vision and Pattern Recognition}, 2019, pp. 10\,609--10\,618.

\bibitem{lee2020making}
M.~A. Lee, Y.~Zhu, P.~Zachares, M.~Tan, K.~Srinivasan, S.~Savarese, L.~Fei-Fei, A.~Garg, and J.~Bohg, ``Making sense of vision and touch: Learning multimodal representations for contact-rich tasks,'' \emph{IEEE Transactions on Robotics}, vol.~36, no.~3, pp. 582--596, 2020.

\bibitem{guzey2024see}
I.~Guzey, Y.~Dai, B.~Evans, S.~Chintala, and L.~Pinto, ``See to touch: Learning tactile dexterity through visual incentives,'' in \emph{2024 IEEE International Conference on Robotics and Automation (ICRA)}.\hskip 1em plus 0.5em minus 0.4em\relax IEEE, 2024, pp. 13\,825--13\,832.

\bibitem{tian2020contrastive}
Y.~Tian, D.~Krishnan, and P.~Isola, ``Contrastive multiview coding,'' in \emph{Computer Vision--ECCV 2020: 16th European Conference, Glasgow, UK, August 23--28, 2020, Proceedings, Part XI 16}.\hskip 1em plus 0.5em minus 0.4em\relax Springer, 2020, pp. 776--794.

\bibitem{lin2024learning}
T.~Lin, Y.~Zhang, Q.~Li, H.~Qi, B.~Yi, S.~Levine, and J.~Malik, ``Learning visuotactile skills with two multifingered hands,'' \emph{arXiv preprint arXiv:2404.16823}, 2024.

\bibitem{miller1999integration}
P.~Miller and P.~Leibowitz, ``Integration of vision, force and tactile sensing for grasping,'' \emph{Int. J. Intell. Mach}, vol.~4, pp. 129--149, 1999.

\bibitem{yu2024octopi}
S.~Yu, K.~Lin, A.~Xiao, J.~Duan, and H.~Soh, ``Octopi: Object property reasoning with large tactile-language models,'' \emph{arXiv preprint arXiv:2405.02794}, 2024.

\bibitem{guzhov2021esresne}
A.~Guzhov, F.~Raue, J.~Hees, and A.~Dengel, ``Esresne (x) t-fbsp: Learning robust time-frequency transformation of audio,'' in \emph{2021 International Joint Conference on Neural Networks (IJCNN)}.\hskip 1em plus 0.5em minus 0.4em\relax IEEE, 2021, pp. 1--8.

\bibitem{guzhov2022audioclip}
------, ``Audioclip: Extending clip to image, text and audio,'' in \emph{ICASSP 2022-2022 IEEE International Conference on Acoustics, Speech and Signal Processing (ICASSP)}.\hskip 1em plus 0.5em minus 0.4em\relax IEEE, 2022, pp. 976--980.

\bibitem{gong2021ast}
Y.~Gong, Y.-A. Chung, and J.~Glass, ``Ast: Audio spectrogram transformer,'' \emph{arXiv preprint arXiv:2104.01778}, 2021.

\bibitem{dosovitskiy2020image}
A.~Dosovitskiy, ``An image is worth 16x16 words: Transformers for image recognition at scale,'' \emph{arXiv preprint arXiv:2010.11929}, 2020.

\bibitem{radford2021learning}
A.~Radford, J.~W. Kim, C.~Hallacy, A.~Ramesh, G.~Goh, S.~Agarwal, G.~Sastry, A.~Askell, P.~Mishkin, J.~Clark \emph{et~al.}, ``Learning transferable visual models from natural language supervision,'' in \emph{International conference on machine learning}.\hskip 1em plus 0.5em minus 0.4em\relax PMLR, 2021, pp. 8748--8763.

\bibitem{schulman2022chatgpt}
J.~Schulman, B.~Zoph, C.~Kim, J.~Hilton, J.~Menick, J.~Weng, J.~F.~C. Uribe, L.~Fedus, L.~Metz, M.~Pokorny \emph{et~al.}, ``Chatgpt: Optimizing language models for dialogue,'' \emph{OpenAI blog}, vol.~2, no.~4, 2022.

\bibitem{perez2018film}
E.~Perez, F.~Strub, H.~De~Vries, V.~Dumoulin, and A.~Courville, ``Film: Visual reasoning with a general conditioning layer,'' in \emph{Proceedings of the AAAI conference on artificial intelligence}, vol.~32, no.~1, 2018.

\bibitem{jang2022bc}
E.~Jang, A.~Irpan, M.~Khansari, D.~Kappler, F.~Ebert, C.~Lynch, S.~Levine, and C.~Finn, ``Bc-z: Zero-shot task generalization with robotic imitation learning,'' in \emph{Conference on Robot Learning}.\hskip 1em plus 0.5em minus 0.4em\relax PMLR, 2022, pp. 991--1002.

\bibitem{palivla}
\BIBentryALTinterwordspacing
K.~Stachowicz, ``{PaliVLA},'' 2024, {GitHub} repository. [Online]. Available: \url{https://github.com/kylestach/bigvision-palivla}
\BIBentrySTDinterwordspacing

\end{thebibliography}

\end{document}